\UseRawInputEncoding
\documentclass[conference]{IEEEtran}
\IEEEoverridecommandlockouts
\usepackage{cite} 
\usepackage{amsmath} 
\usepackage{amssymb} 
\usepackage{amsfonts} 
\usepackage{algorithmic} 
\usepackage{graphicx} 
\usepackage{textcomp} 
\usepackage{xcolor} 
\usepackage{booktabs} 
\usepackage{adjustbox}

\usepackage{makecell}
    
\begin{document}

\title{SPFFNet: Strip Perception and Feature Fusion Spatial Pyramid Pooling for Fabric Defect Detection\\}

\author{
\IEEEauthorblockN{1\textsuperscript{st} Peizhe Zhao}
\IEEEauthorblockA{
\textit{Waterford Institute} \\
\textit{South East Technological University} \\
Waterford, Ireland \\
asher.zhao.ai@outlook.com}
\and
\IEEEauthorblockN{2\textsuperscript{nd} Shunbo Jia}
\IEEEauthorblockA{
\textit{Faculty of Innovation Engineering} \\
\textit{Macau University of Science and Technology}\\
Macau, China \\
2240003657@student.must.edu.mo}
}

\maketitle

\begin{abstract}
Defect detection in fabrics is critical for quality control, yet existing methods often struggle with complex backgrounds and shape-specific defects. In this paper we propose SPFFNet, an improved fabric defect detection model based on the YOLOv11 framework. To enhance the detection of strip defects, we introduce a Strip Perception Module (SPM) that improves feature capture through multi-scale convolution. We further enhance the spatial pyramid pooling fast (SPPF) by integrating a squeeze-and-excitation mechanism, resulting in the SE-SPPF module, which better integrates spatial and channel information for more effective defect feature extraction. Additionally, we propose a novel focal enhanced complete intersection over union (FECIoU) metric with adaptive weights, addressing scale differences and class imbalance by adjusting the weights of hard-to-detect instances through focal loss. Experimental results demonstrate that our model achieves a 0.8-8.1\% improvement in mean average precision (mAP) on the Tianchi dataset and a 1.6-13.2\% improvement on our custom dataset, outperforming other state-of-the-art methods.
\end{abstract}

\begin{figure*}[t]
    \centering
    \includegraphics[width=\textwidth]{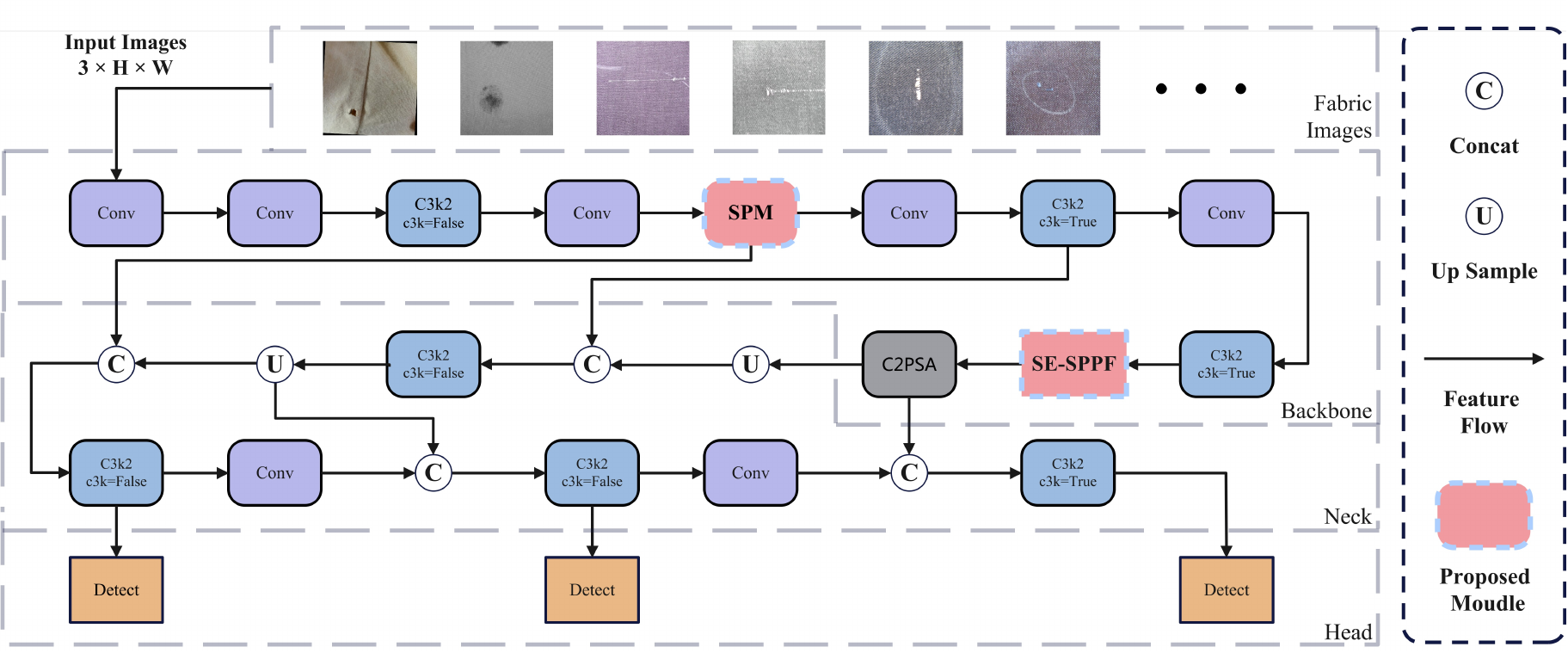}
    \caption{Network structure of the proposed method}
\label{fig1}
\end{figure*}

\section{Introduction}

Traditional fabric defect detection~\cite{weng_2024_enhancing, li_2024_lrfpn, qiao_2022_a} relies heavily on visual inspection by human experts, a process that is time-consuming, labor-intensive, and prone to errors, particularly when defects are small or contrast is low. This method often produces subjective and difficult-to-quantify results, leading to high defect rates and unreliable assessments. As a result, computer vision-based defect detection algorithms, have begun to emerge and develop. However, general object detection algorithms struggle with the complex backgrounds of fabric defects and their varied aspect ratios. Thus, adapting to the large-scale variations of fabric defects and distinguishing complex backgrounds are key challenges in improving the performance of fabric defect detection.

Modern fabric defect detection algorithms are generally divided into two categories: two-stage and single-stage methods. The two-stage method, such as Zhao \textit{et al.} ~\cite{jia_2022_fabric} proposed a transfer learning-based Faster Region-based Convolutional Neural Network (Faster R-CNN), enhancing fabric defect detection accuracy via a cascaded module. However, it faces challenges in training efficiency, computational cost, and generalizability to complex textures and diverse defects. 

The single-stage method, derived from the YOLO~\cite{redmon_2016_you} framework, has shown promise. For example, the enhanced YOLOv3~\cite{redmon_2018_yolov3} model~\cite{jing_2020_fabric} improves detection through an attention mechanism and negative sample weighting but remains insufficient for accurately detecting complex defect types. The YOLOv5~\cite{jocher_2020_ultralyticsyolov5} algorithm~\cite{liu_2023_an} enhances feature representation by combining adaptive pooling with an attention module~\cite{zhao2024athlete, luo2025tennis} and optimizing the loss function. However, its accuracy remains limited in handling specific defect types and complex scenarios.

In response to the aforementioned challenges, we propose SPFFNet, a novel architecture built upon the YOLOv11 framework~\cite{jocher_2023_yolov8}. While preserving the inference efficiency characteristic of single-stage detectors, SPFFNet introduces a Strip Perception Module (SPM) that leverages multi-scale convolution to substantially enhance the network’s capability for fine-grained feature extraction and representation of strip defects. To further improve discrimination between complex background textures and subtle defect regions, we design an enhanced Squeeze-and-Excitation Spatial Pyramid Pooling Fast (SE-SPPF) module, which effectively integrates spatial and channel-wise information to achieve more comprehensive contextual understanding. Moreover, to address the substantial variation in bounding-box scales across different defect categories, we propose a Focal Enhanced Complete Intersection over Union (FECIoU) metric. This metric dynamically reweights difficult-to-detect samples, thereby improving robustness and adaptability to targets with extreme aspect ratios.The main contributions of this work are summarized as follows:
\begin{itemize}
    \item[\textcolor{black}{$\bullet$}] A multi-scale convolutional SPM is introduced into the YOLOv11 backbone to improve feature capture and extraction for strip defects.
    \item[\textcolor{black}{$\bullet$}] SE-SPPF is proposed to enhance the model's ability to distinguish complex backgrounds and targets by combining weighted channel maps with spatial pyramid pooling.
    \item[\textcolor{black}{$\bullet$}] We propose FECIoU, which incorporates a focal weighting mechanism to reduce the impact of scale variations in fabric defects.
    \item[\textcolor{black}{$\bullet$}] We have collected, organized, and annotated a fabric defect dataset consisting of 8,645 samples.
\end{itemize}

\section{Related Work}\label{sec:rw}

\subsection{Object Detection}
Object detection has long been an active research area, with numerous methods developed to enhance detection accuracy and efficiency. Early approaches such as R-CNN~\cite{girshick_2014_rich} combined region proposals with CNN-based feature extraction, while Fast R-CNN~\cite{girshick_2015_fast} and Faster R-CNN~\cite{ren_2017_faster} improved both speed and accuracy through shared convolutional features and the introduction of the Region Proposal Network (RPN). Subsequently, one-stage detectors like YOLO~\cite{redmon_2016_you} and SSD~\cite{liu_2016_ssd} achieved real-time object detection by formulating detection as a single regression problem. Despite these advances, challenges remain in detecting small objects and handling diverse object scales and shapes.

In object detection, the loss function quantifies the difference between predicted and ground truth bounding boxes. Intersection over Union (IoU)~\cite{jiang_2018_acquisition} is commonly used to measure this overlap. The IoU loss encourages the model to align predicted boxes with ground truth. The Generalized IoU (GIoU)~\cite{rezatofighi_2019_generalized} extends IoU by addressing scale and offset mismatches, providing more reliable localization, but it can be ineffective for boxes with significant overlap.
Distance IoU (DIoU)~\cite{zheng_2019_distanceiou} refines GIoU by incorporating centroid distance, improving localization accuracy. However, DIoU does not account for size variations between objects. Complete IoU (CIoU)~\cite{zheng_2019_distanceiou} incorporates centroid distance, overlap area, and angular difference, making it more effective for rotated boxes. However, for fabric defect detection, where target aspect ratios vary significantly, basic IoU can lead to errors. To address this, we propose an improved version of CIoU (FECIoU), which adjusts for scale differences and enhances detection accuracy for targets with varying aspect ratios.

\subsection{Fabric Defect Detection Algorithms}
Modern fabric defect detection methods are mainly divided into two categories: two-stage and single-stage approaches. The two-stage methods, such as Zhao \textit{et al.} ~\cite{jia_2022_fabric} proposed method, which integrates transfer learning and an improved Faster R-CNN with Residual Network with 50 layers (ResNet50), Feature Pyramid Network (FPN), Region of Interest Align (ROI Align), significantly enhances detection accuracy and robustness for fabric defect detection by employing a cascaded module to refine localization precision. 
Single-stage methods, particularly those based on the YOLO framework, have gained popularity. Enhanced YOLOv3~\cite{redmon_2018_yolov3} improves fabric defect detection by adding an attention mechanism and negative sample weighting~\cite{jing_2020_fabric}. While effective, it still underperforms in detecting complex defects. YOLOv5~\cite{jocher_2020_ultralyticsyolov5} improves feature representation through adaptive pooling and an attention module~\cite{liu_2023_an}, but faces challenges in complex scenarios. To address these issues, we propose an improved YOLOv4-based model with a Strip Perception Module (SPM) that enhances feature extraction for strip defects, retaining the speed advantage of single-stage detection.

\subsection{Attention Mechanism}
The attention mechanism~\cite{shen_2023_pedestrianspecific, shen_2023_triplet} enhances model performance by focusing on relevant spatial, channel, or hybrid features. Spatial attention methods like SAM~\cite{zhu_2019_an} and RANet~\cite{shao_2023_ranet} prioritize key regions in the spatial domain, improving the capture of spatial dependencies. RANet uses a relation module to model feature interactions, leveraging attention or graph convolutions.
Channel attention, exemplified by SENets~\cite{hu_2018_squeezeandexcitation}, introduces a squeeze-and-excitation (SE) block that reweights feature channels to highlight important features. This mechanism improves representational power without significantly increasing computational cost. For fabric defect detection, we propose an enhanced spatial pyramid pooling fast (SE-SPPF) that integrates SENetv2~\cite{narayanan_2023_senetv2} for better multi-scale feature fusion, addressing the complexity and variation of defect shapes.

\begin{figure}[t]
    \centering
    \includegraphics[width=\linewidth]{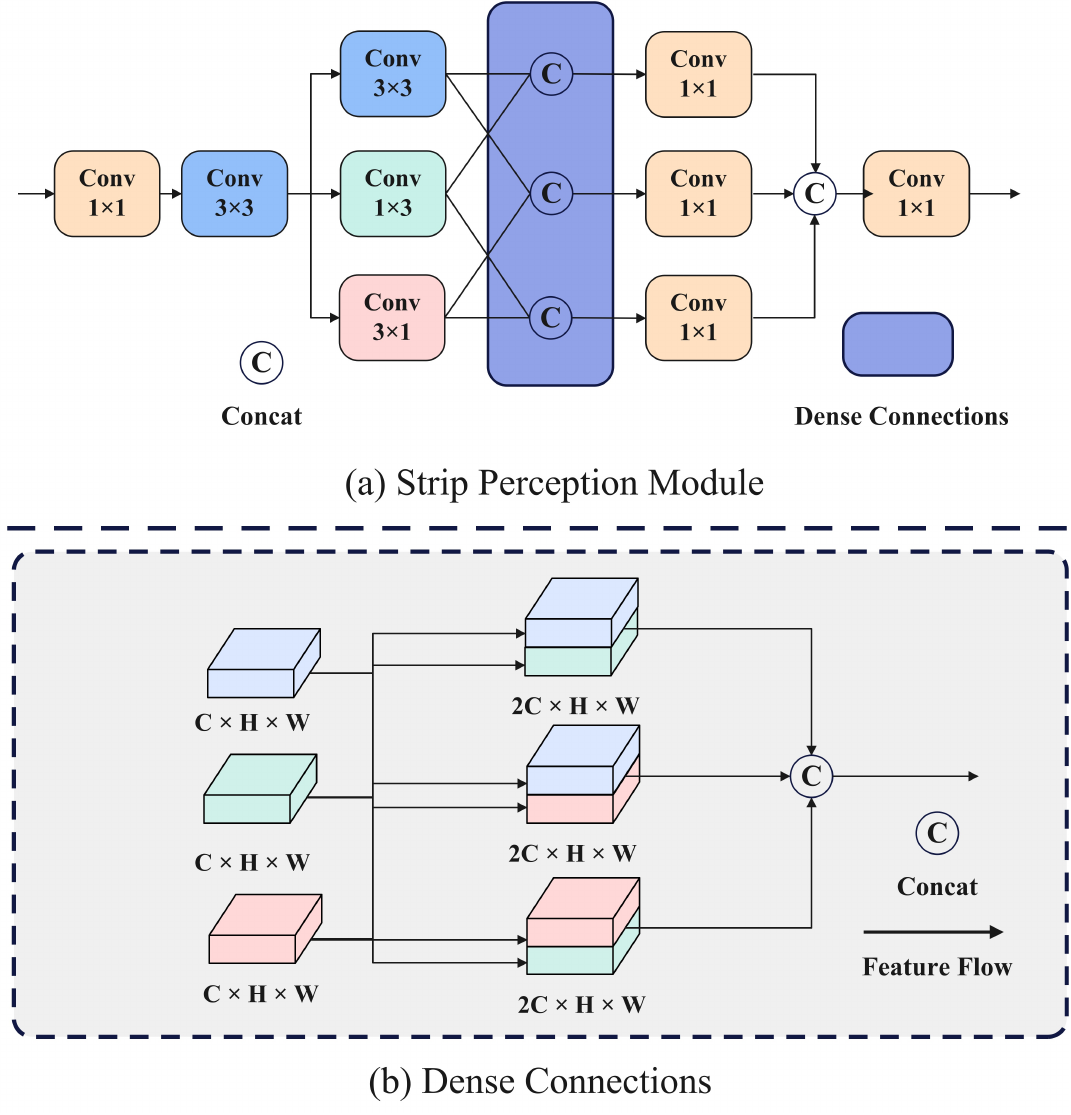}
    \caption{(a) illustrates the overall architecture of the Strip Perception Module, while (b) presentsdetailed fusion operations within the Dense Connections component. The Dense Connectionsfacilitate comprehensive integration of diverse strip-wise features.}
\label{fig2}
\end{figure}

\section{Proposed Method}\label{sec:method} 

\subsection{Overview}
This paper presents a fabric defect detection method based on YOLOv11, addressing the challenges of complex defect shapes and the need for high detection accuracy and real-time performance. The proposed method incorporates a strip perception module (SPM) and a squeeze-and-excitation spatial pyramid pooling fast (SE-SPPF). As shown in Fig.\ref{fig1}, this approach enhances YOLOv11 by maintaining high detection accuracy while meeting real-time constraints, achieving significant improvements in fabric defect detection.
The SPM leverages strip convolution to extract strip defect features through intensive interactions with convolutions of various shapes, improving the model's precision in detecting and positioning strip defects. To enhance background discrimination and texture information extraction, the spatial pyramid pooling is re-designed as SE-SPPF, combining the channel attention mechanism of SENetv2. This module optimally utilizes both channel and spatial information to refine background discrimination and defect feature extraction. Additionally, a novel loss function, focal enhanced complete intersection over union (FECIoU), is introduced to address the issue of large-scale variations in target boxes. FECIoU assigns higher weights to samples with lower IoU, ensuring the model focuses on these challenging samples during training, thus improving detection efficiency and accuracy.

\subsection{Strip Perception Module}  
In the task of fabric defect detection, the complex shape and large size variation of defect features affect the accuracy of detection. Multi-scale convolution can effectively capture features at different scales in the feature map, especially when facing long strip-shaped defects that occur frequently in fabric operations. Multi-scale convolution can more effectively extract defect features. The specific design is shown in Fig.\ref{fig2}.
This paper proposes SPM. First, two convolution blocks of 1x1 and 3x3 are used to minimize the number of channels, and then multi-scale (1x3, 3x1, 3x3) convolution operations are performed using branch parallelism. The resulting feature maps are densely stacked using concat, and then a 1x1 convolution kernel is used to extract important features from the convolutions of different scales. Finally, a residual structure was introduced to improve the stability and effectiveness of training. While maintaining the depth of the network, information transmission and gradient flow are ensured. In summary, SPM can effectively extract the features of strip defects and improve the accuracy of the model. 

\begin{figure}[t]
    \centering
    \includegraphics[width=\linewidth]{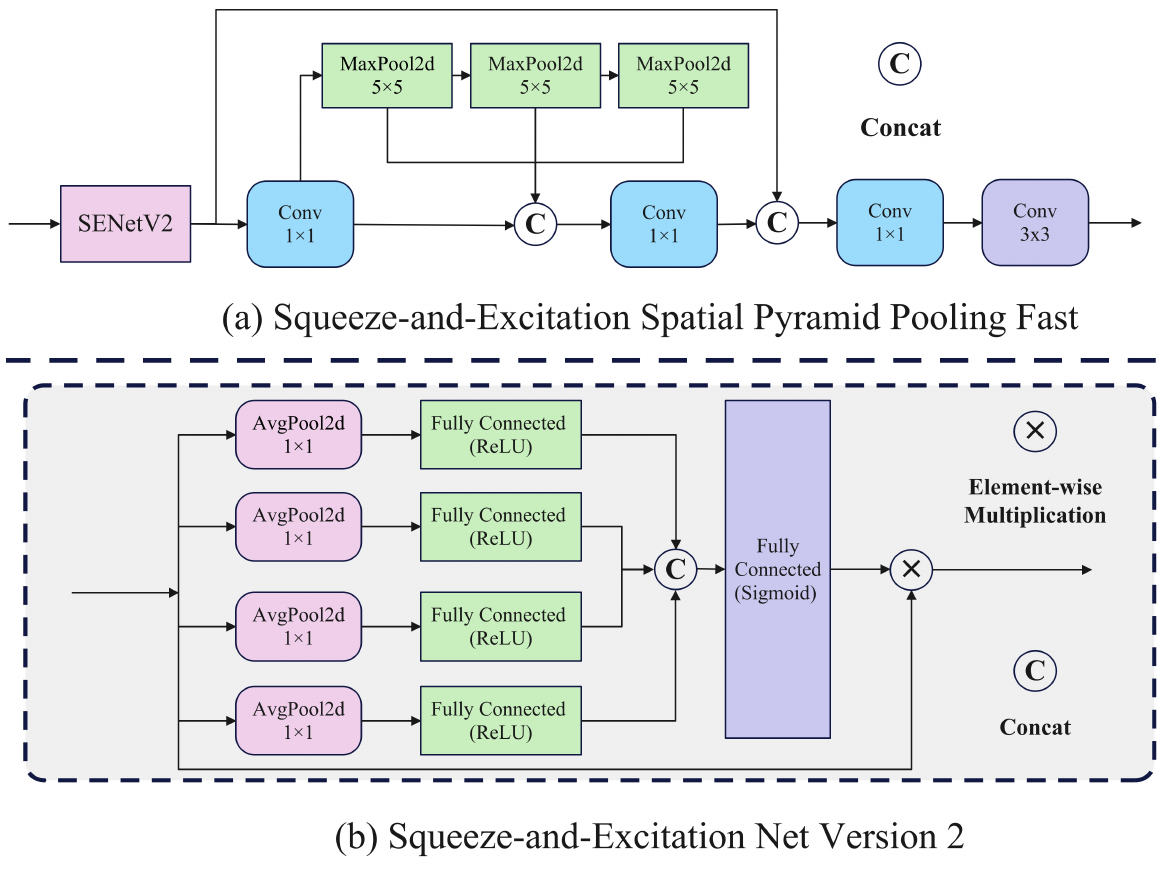}
    \caption{(a) illustrates the overall architecture of the Squeeze-and-Excitation Spatial Pyramid Pooling Fast (SE-SPPF), while (b) provides a detailed breakdown of the SENetV2 module's processing pipeline. The SENetV2 module effectively harmonizes multi-scale features to extract the most discriminative characteristics for defect detection.}
\label{fig3}
\end{figure}

\begin{table*}[t]
\centering
\caption{Comparison of the performance of the proposed improved model with multiple SOTA on the Tianchi dataset}
\label{tab:tianchi}
\renewcommand{\arraystretch}{1.5} 
\begin{adjustbox}{max width=\textwidth}
\begin{tabular}{lccccccccccc}
\toprule
&\multicolumn{9}{l}{\textbf{mAP@0.5/\%}}& & \\
\cmidrule(lr){2-10}
\textbf{Method} & \textbf{Knot} & \textbf{Triple Wire} & \textbf{Coarse Pick} & \textbf{Broken Spandex} & \textbf{Warp Knot}& \textbf{Weft Shrink} & \textbf{Hole} & \textbf{Stain} & \textbf{All} & \textbf{GFLOPs}& \textbf{Params} \\
\midrule
YOLOv5~\cite{jocher_2020_ultralyticsyolov5} & 61.6 & 77.2 & 59.4 & \textbf{76.8} & 45.8 & \textbf{46.9} & \textbf{83.7} & 43.5 & 61.9 & 5.8 & 2183224\\
YOLOv6~\cite{li_2022_yolov6} & 58.1 & 78 & 52.8 & 68.1 & 47.1 & 30.8 & 82.3 & 44.5 & 57.7 & 11.5 & 4155816\\
YOLOv8~\cite{jocher_2023_yolov8} & \textbf{65.9} & 78.8 & 60.5 & 76.3 & 51.3 & 40.1 & 81.6 & 59.9 & 64.3 & 6.8 & 2685928\\
YOLOv9t~\cite{wang_2024_yolov9} & 65.4 & 80.4 & 59.8 & 71.8 & \textbf{52.6} & \textbf{46.9} & 83.3 & 62.9 & 65.4 & 6.4 & 1731384\\
YOLOv9s~\cite{wang_2024_yolov9} & 66.0& \textbf{82.0}& 54.3 & 76.6 & 54.4 & 46.7 & 79.7 & 64.4 & 65.5 & 22.1 & 6196744\\
YOLOv10n~\cite{wang_2024_yolov10} & 59.3 & 77.4 & 57.7 & 69.4 & 41.5 & 39.2 & 81.7 & 57.7 & 60.5 & 8.2 & 2697536\\
YOLOv11n~\cite{jocher_2023_yolov8} & 64.4 & 80.0& \textbf{64.3} & 76.1 & 48.1 & 43.7 & 80.5 & 62.9 & 65.0& 6.3 & 2583712\\
\toprule
SPFFNet (Ours) & 64.5 & 80.5 & 63.5 & 74.6 & 49.0& 43.9 & \textbf{83.7} & \textbf{66.4} & \textbf{65.8} & 6.8 & 2858951\\
\bottomrule
\end{tabular}
\end{adjustbox}
\end{table*}

\subsection{Squeeze and Excitation Spatial Pyramid Pooling Fast } 
Fabric defects usually exhibit multiple features. In order to eliminate some noise, make the features more robust, and help the model better capture the overall structure and texture of the image, SE-SPPF introduces SENetv2 to more reasonably assign weights to each channel. Combined with the multi-scale fusion in SPPF space, it strengthens the model's ability to extract features from both spatial and channel perspectives. The specific design is shown in Fig.\ref{fig3}.
This paper proposes SE-SPPF. First, the feature map is weighted by SENetv2 to the channel, and then the channel number is adjusted using a $1\times1$ convolution and input to SPPF. The four feature maps of different scales obtained by SPPF are concatenated using a residual structure and the weighted feature map Concat after feature extraction using a $1\times1$ convolution. Finally, features are further extracted using two convolutions of $1\times1$ and $3\times3$.

\subsection{Focal Enhanced Complete Intersection over Union}  
The span of the defect detection box for different types of fabric defects is very large, especially for defects that appear in the form of stripes, which are several times or even more than the length and width of most target detection objects. Therefore, this paper proposes FECIoU, which uses a focal weight mechanism to make the model pay more attention to difficult-to-detect objects during training.
Equation \ref{equation1} is the formula for FECIoU, where \((1 - IoU) ^ \gamma\)is the weight value for CIoU and \(\gamma\) is a manually set parameter. In Equation \ref{equation2} ,\(\rho^2(b, b^g)\)is the squared Euclidean distance between the centers of the predicted and ground truth boxes, calculated as shown in Equation \ref{equation3}, and \( c \) is the diagonal length of the minimum bounding box. \(\alpha v\) is a penalty term for the aspect ratio difference, and the specific calculation method is shown in Equations \ref{equation4} and \ref{equation5} . \( w^g, h^g, w, \) and \( h \) are the width and height of the predicted frame and the actual frame, respectively.

\begin{equation}
\text{FECIoU} = (1 - \text{IoU})^\gamma \cdot \left( \text{IoU} - \frac{\rho^2(\mathbf{b}, \mathbf{b}^g)}{c^2} - \alpha v \right),
\label{equation1}
\end{equation}

\begin{equation}
\text{CIoU} = \text{IoU} - \frac{\rho^2(\mathbf{b}, \mathbf{b}^g)}{c^2} - \alpha v,
\label{equation2}
\end{equation}

\begin{equation}
\rho^2(\mathbf{b}, \mathbf{b}^g) = (x_b - x_{b^g})^2 + (y_b - y_{b^g})^2,
\label{equation3}
\end{equation}

\begin{equation}
v = \frac{4}{\pi^2} \left( \arctan \frac{w^g}{h^g} - \arctan \frac{w}{h} \right)^2,
\label{equation4}
\end{equation}

\begin{equation}
\quad \alpha = \frac{v}{(1 - \text{IoU}) + v}.
\label{equation5}
\end{equation}

\section{Experiment and Analysis}\label{sec:exp} 
\subsection{Datasets}
\subsubsection{Tianchi fabric dataset}
Tianchi fabric dataset~\cite{tianchi_2020_smart}, provided by Alibaba's Tianchi platform, is a significant resource for fabric defect detection research. It comprises high-resolution fabric images with detailed annotations of various defect types, such as holes, stains, wrinkles, color shades, and missing threads. The dataset, consisting of thousands to tens of thousands of images, is designed to facilitate the development and validation of defect detection algorithms and automated quality inspection systems in the fabric industry. 

\begin{table*}[t]
\centering
\caption{Comparison of the performance of the proposed improved model with multiple SOTA on the produced dataset}
\label{tab:produced}
\renewcommand{\arraystretch}{1.5} 
\begin{adjustbox}{max width=\textwidth}
\begin{tabular}{lcccccccc}
\toprule
&\multicolumn{6}{l}{\textbf{mAP@0.5/\%}}& & \\
\cmidrule(lr){2-7}
\textbf{Method} & \textbf{Missing Stitches} & \textbf{Broken Holes} & \textbf{Stain} & \textbf{Broken Seam} & \textbf{Broken Stitches} & \textbf{All} & \textbf{GFLOPs}& \textbf{Params} \\
\toprule
YOLOv5~\cite{jocher_2020_ultralyticsyolov5} & 85.4 & 73.4 & 99.5 & 80.2 & 75.9 & 82.9 & 5.8 & 2182639\\
YOLOv6~\cite{li_2022_yolov6} & 83.0& 68.9 & 99.5 & 80.2 & 55.5 & 77.4 & 11.5 & 4155519\\
YOLOv8~\cite{jocher_2023_yolov8} & 93.9 & 78.2 & 99.5 & 82.0& 88.1 & 88.3 & 6.8 & 2685343\\
YOLOv9t~\cite{wang_2024_yolov9} & 89.1 & 76.3 & 99.5 & 82.1 & 85.8 & 86.5 & 6.4 & 1730799\\
YOLOv9s~\cite{wang_2024_yolov9} & 91.7 & 80.2 & 99.5 & 81.2 & 91.8 & 88.9 & 22.1 & 6195583\\
YOLOv10n~\cite{wang_2024_yolov10} & 89.5 & 76.8 & 99.5 & 78.6 & 85.9 & 86.1 & 8.3 & 2696336 \\
YOLOv11n~\cite{jocher_2023_yolov8} & 93.1 & 79.4 & 99.5 & \textbf{83.8}& 89.3 & 89.0& 6.3 & 2583127 \\
\midrule
SPFFNet (Ours) & \textbf{95.3}& \textbf{83.5}& \textbf{99.5}& 81.1 & \textbf{93.5}& \textbf{90.6}& 6.8 & 2858951 \\
\bottomrule
\end{tabular}
\end{adjustbox}
\end{table*}

\subsubsection{Produced dataset}
This dataset was collected and labeled and organized by us. The data mainly comes from the workshop of a fabric factory in Jiangsu Province and public images that can be collected on the Internet. After our collection and organization, the final dataset contains a total of 8,645 fabric defect images, which are classified into five types of defects that are most commonly found in the fabric process: missing stitches, broken holes, stain, broken seam, and broken stitches. In addition, this paper also uses some image data enhancement methods, such as rotation, translation, scaling, and flipping, to expand the dataset and generate more samples, thereby improving the generalization ability of the model and reducing the risk of over-fitting.

\subsection{Implementation Details} 
All experiments were conducted on an NVIDIA RTX 4090D GPU, with the YOLO series models configured to use their standard (“normal”) size variants. The models were trained with a batch size of 32 and an input resolution of 640 × 640. Given the large scale of the dataset and the potential presence of noisy samples, Stochastic Gradient Descent (SGD) was adopted as the optimizer to enhance convergence stability and mitigate the risk of local minima, with an initial learning rate of 0.01 and momentum of 0.937. To ensure fair comparison across models of different sizes, the early stopping patience was uniformly set to 20 epochs, allowing training to continue for up to 20 epochs without improvement in validation accuracy before termination.

\subsection{Comparison with State-of-the-art Methods} 
We compare the proposed SPFFNet with six state-of-the-art object detection models, including YOLOv5 \cite{jocher_2020_ultralyticsyolov5}, YOLOv6 \cite{li_2022_yolov6}, YOLOv8 \cite{jocher_2023_yolov8}, YOLOv9-t \cite{wang_2024_yolov9}, YOLOv9-s \cite{wang_2024_yolov9}, and YOLOv10-n \cite{wang_2024_yolov10}, to comprehensively assess detection accuracy and efficiency under consistent experimental conditions.

\subsubsection{Comparisons on Tianchi fabric dataset}
Table \ref{tab:tianchi} shows a comparison of the performance of the proposed improved model with multiple state-of-the-art algorithms on the Tianchi dataset. It can be seen that the model proposed in this paper achieved the highest mAP (i.e., 65.8\%).

The mAP of the improved model in each defect category performed well, which shows that the proposed SE-SPPF module fully integrates important defect information from both spatial and channel perspectives, helping the model find key features.

\subsubsection{Comparisons on produced dataset}
Table \ref{tab:produced} shows a comparison of the performance of the proposed improved model with multiple state-of-the-art algorithms on the dataset we created. It can be seen that the model proposed in this paper achieves the highest mAP (i.e. 90.6\%) without significantly increasing the computational cost and model size. Among them, the mAP for the detection of the two strip defects missing stitches and broken stitch is the highest among all methods. This shows that the multi-scale convolution SPM plays a key role in the detection of strip defects, which improves the detection ability of the model.

\begin{table}[t]
\centering
\caption{Results of ablation experiments on produced datasets}
\renewcommand{\arraystretch}{1.5} 
\begin{tabular}{cccccc}
\toprule
\textbf{SPM} & \textbf{SE-SPPF} & \textbf{FECIoU} & \textbf{mAP@0.5/\%} & \textbf{GFLOPs} & \textbf{Params} \\ \midrule
- & - & - & 89& 6.3& 2583127\\
\checkmark & - & - & 89.6& 6.6& 2613063\\
- & \checkmark & - & 89.6& 6.6& 2894679\\
\checkmark & \checkmark & - & 90.3& 6.8& 2858951\\
\checkmark & \checkmark & \checkmark & 90.6& 6.8& 2858951 \\ \bottomrule
\end{tabular}
\label{tab:producedab}
\end{table}

\subsection{Ablation Studies and Analysis}  
Tables~\ref{tab:tianchi} and~\ref{tab:produced} demonstrate that the proposed model consistently outperforms several state-of-the-art single-stage detectors. To further substantiate its effectiveness, an ablation study was conducted on the custom dataset (Table~\ref{tab:producedab}). Integrating the SPM, SE-SPPF, and FECIoU modules yields the best performance, achieving 90.6\% mAP with only a marginal increase in computation (6.8 vs. 6.3 GFLOPs). Specifically, SPM enhances strip-oriented feature perception, while SE-SPPF strengthens spatial–channel interactions; both contribute notable accuracy improvements with negligible overhead. Their synergistic combination demonstrates strong complementarity, and the inclusion of FECIoU further refines localization, resulting in the highest overall accuracy. These findings confirm that the proposed components effectively boost detection capability while maintaining computational efficiency, underscoring the robustness and practicality of SPFFNet for real-world fabric defect detection.

\begin{figure}[t]
    \centering
    \includegraphics[width=1\linewidth]{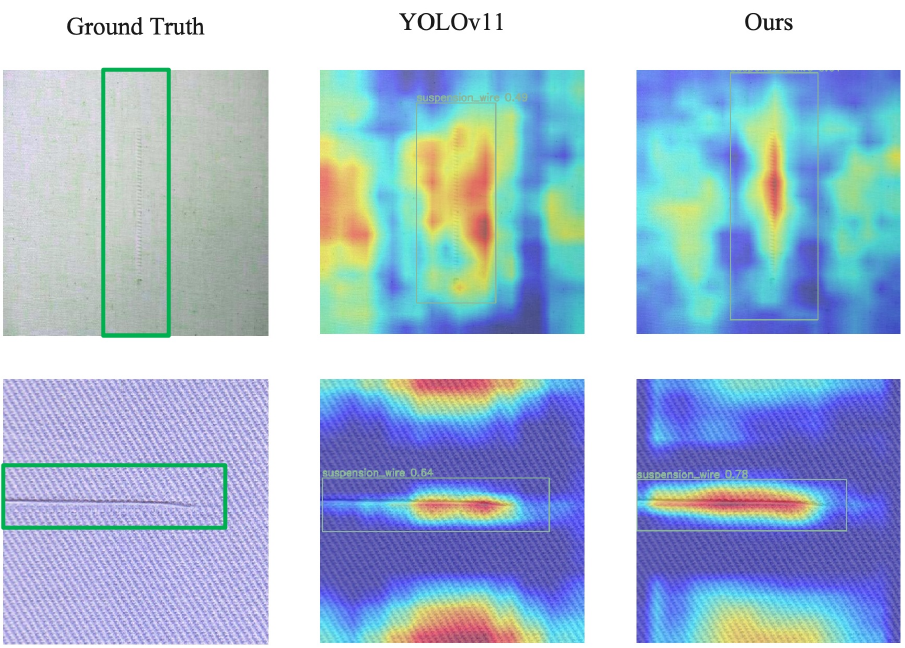}
    \caption{Comparison visualized by heat maps}
\label{fig4}
\end{figure}

\subsection{Visualization}  
As shown in Fig.\ref{fig4}, the heat maps after the spatial pyramid pooling layer of the baseline model and the improved model proposed in this paper are shown respectively. It can be intuitively seen that the improved model proposed in this paper is more accurate than the baseline model in determining the most important region for prediction, and the coverage completely includes the defective parts of this fabric. This shows that the SPM module accurately extracts the important features of the strip defects, and SE-SPPF allows the model to accurately distinguish between the background and defects, which in turn allows the model to more accurately determine the most important region for judgment. The visualization results of the heat map once again verify the effectiveness of the structure proposed in this paper.

\section{Conclusion}\label{sec:con} 
In this paper, we propose SPFFNet, an enhanced fabric defect detection framework built upon YOLOv11, which integrates the Strip Perception Module (SPM), Squeeze-and-Excitation Spatial Pyramid Pooling Fast (SE-SPPF), and Focal Enhanced Complete IoU (FECIoU) loss to improve feature representation, background discrimination, and localization precision. Extensive experiments on the Tianchi and custom datasets demonstrate that SPFFNet achieves consistent gains over state-of-the-art approaches, confirming its effectiveness for complex industrial inspection scenarios. 

However, the current model is still limited by its reliance on RGB imagery and a relatively narrow range of defect categories, which may restrict its generalization to diverse textile materials and illumination conditions. Future work will focus on enhancing the model’s robustness to color variations and unseen defect patterns through cross-domain learning and spectral feature integration, as well as improving its efficiency and adaptability for real-time deployment in large-scale manufacturing environments.

\bibliographystyle{IEEEtran}
\bibliography{ref}
\end{document}